\documentclass{article}

\usepackage[final,main]{neurips_2025}
\usepackage{natbib} 
\usepackage{multirow}
\usepackage{booktabs}
\usepackage{enumitem}

\usepackage[utf8]{inputenc} 
\usepackage[T1]{fontenc}    
\usepackage{hyperref}       
\usepackage{url}            
\usepackage{booktabs}       
\usepackage{amsfonts}       
\usepackage{nicefrac}       
\usepackage{microtype}      
\usepackage{xcolor}         
\usepackage{graphicx}
\usepackage{amsmath}
\usepackage{float} 

\title{Discovering Transformer Circuits via a Hybrid Attribution and Pruning Framework}

\author{
  Hao Gu\thanks{Co-first authors} \quad
  Vibhas Nair\footnotemark[1] \quad
  Amrithaa Ashok Kumar \quad
  Jayvart Sharma \quad
  Ryan Lagasse\thanks{Correspondence: \texttt{ryan@algoverseairesearch.org}} \\
  Algoverse AI Research \\
  \texttt{ryan@algoverseairesearch.org}
}

\begin{document}

\maketitle

\begin{abstract}
\label{abs}
  Interpreting language models often involves circuit analysis, which aims to identify sparse subnetworks, or \textit{circuits}, that accomplish specific tasks. Existing circuit discovery algorithms face a fundamental trade-off: attribution patching is fast but unfaithful to the full model, while edge pruning is faithful but computationally expensive. This research proposes a hybrid attribution and pruning (HAP) framework that uses attribution patching to identify a high-potential subgraph, then applies edge pruning to extract a faithful circuit from it. We show that HAP is 46\% faster than baseline algorithms without sacrificing circuit faithfulness. Furthermore, we present a case study on the Indirect Object Identification task, showing that our method preserves cooperative circuit components (e.g. S-inhibition heads) that attribution patching methods prune at high sparsity. Our results show that HAP could be an effective approach for improving the scalability of mechanistic interpretability research to larger models\footnote{Our code is available at: \url{https://anonymous.4open.science/r/HAP-circuit-discovery}}.
\end{abstract}

\section{Introduction}
\label{intro}
Large language models (LLMs) are increasingly being deployed in high-stakes settings, motivating the need to uncover their "black-box" \cite{Alishahi_Chrupała_Linzen_2019} nature and understand how they "think." \cite{hubinger2020overview11proposalsbuilding,zhang_a_survey} This is a key goal of mechanistic interpretability, a field focused on understanding transformer \cite{vaswani2023attentionneed} model behavior by analyzing the interactions between subnetworks of attention heads and multi-layer perceptrons (MLPs)\cite{vig2020causalmediationanalysisinterpreting,sharkey2025openproblemsmechanisticinterpretability}. The most common approach to mechanistic interpretability is through circuit analysis, which identifies sparse subnetworks, or "circuits", responsible for specific behaviors \cite{olah_zoom_2020, olah_mechanistic_nodate,erdogan_automated_nodate}. Manual circuit discovery methods, such as that proposed by \cite{wang_interpretability_2022}, have largely been replaced by automated approaches like Automated Circuit DisCovery (ACDC) \cite{conmy_towards_2023}, which uses a greedy search algorithm to ablate edges one by one. 

To address the computational cost of ACDC, faster algorithms such as Edge Attribution Patching (EAP) \cite{syed_attribution_2023} and Edge Pruning (EP) \cite{bhaskar_finding_2025} have been proposed. However, existing circuit discovery algorithms struggle to scale with larger models without sacrificing performance \cite{hanna_have_2024,hsu_efficient_2025,zhang_eap-gp_2025}. EAP uses a first-order Taylor series approximation to ablate all edges simultaneously. Although faster than ACDC, the first-order approximations show low faithfulness to the full model. Conversely, EP efficiently applies a gradient-based pruning algorithm to discover circuits in parallel. Despite scaling well to larger models while maintaining exceptional circuit faithfulness, EP requires significant compute power. 

This research proposes a novel Hybrid Attribution and Pruning (HAP) framework to enhance the scalability and maintain the faithfulness of discovered circuits. We leverage EAP to quickly filter out the majority of unimportant edges. This EAP-identified subgraph gives a narrowed search space for EP to find faithful circuits.
In summary, our main contributions are the following:
\begin{enumerate}[itemsep=0pt, topsep=0pt]
  \item We propose a novel framework (HAP) that improves efficiency and preserves the faithfulness of discovered circuits.
  \item We show that HAP matches or outperforms existing methods in efficiency and faithfulness.
  \item We demonstrate in an IOI case study that HAP finds the often-missed S-Inhibition heads, preserving the quality of discovered circuits.
\end{enumerate}
\section{Related Works}
\label{sec:related-work}
{\bf Automated Circuit Discovery Algorithms} such as ACDC construct computational graphs where nodes represent model components and edges represent information flow \cite{conmy_towards_2023}. ACDC recursively applies activation patching—replacing activations with those from “corrupted” examples—removing edges that do not degrade task metric performance \cite{syed_attribution_2023}. This greedy search can rediscover known circuits, but it is computationally expensive for larger models or datasets due to the requirement for many forward passes, with scalability limited by the number of edges evaluated \cite{conmy_towards_2023}.

{\bf Edge Pruning and Optimization-based Methods} frame circuit discovery as a gradient-based optimization problem, where edges between components of a model's computational graph are pruned using binary masks over edges \cite{bhaskar_finding_2025}. This method allows for finer-grained and more faithful recovery of causal pathways, but requires architectural modifications and additional memory for scalability. EP can parallelize training across multiple GPUs, which enables EP to scale to large models (e.g., CodeLlama-13B) and complex datasets, recovering circuits that are both smaller and more interpretable than those produced by prior methods \cite{bhaskar_finding_2025}. 

{\bf Attribution and Gradient-based Approximations}, like EAP, propose gradient-based, first-order approximations to activation patching \cite{syed_attribution_2023}, enabling simultaneous computation of edge importance scores with one backward and two forward passes. EAP efficiently identifies circuits that align closely with those found by ACDC, as measured by ROC/AUC when compared to manually curated circuit ground-truths, but can miss critical component interactions due to its linear approximation and reduced faithfulness \cite{bhaskar_finding_2025}.   

\section{Methods}
\label{methods}
The HAP framework operates by leveraging EAP to perform a global search, quickly removing low-importance edges to isolate higher-importance edges. This EAP-identified subgraph gives a narrowed search space for the precise pruning algorithm, EP.

\subsection{Step 1: Computational Graph Construction}
\label{graph}
We start by representing our model as a computational graph following the convention of \cite{bhaskar_finding_2025}, where components of the Transformer architecture, namely attention layers and MLPs, are the nodes and the edges between any two nodes represent the connection between the output of one node to the input of the other node. The full model, in our case GPT-2 Small (from \cite{Radford2019LanguageMA}), can be represented at this granularity, and a circuit is a computational subgraph consisting of a set of edges that describe the full model’s behavior on a particular task (see Section \ref{sec:ioi}). 

\subsection{Step 2: Edge Attribution Patching}
\label{eap}
We then use Edge Attribution Patching to quickly get absolute attribution scores that measure the importance of all edges in the computational graph using:  
\begin{equation} \label{strain}
L(\mathbf{x} \mid e_{\text{ablated}}) - L(\mathbf{x}) 
\;\approx\; 
\left( e_{\text{clean}} - e_{\text{ablated}} \right)^{\top} 
\frac{\partial L(\mathbf{x} \mid e_{\text{clean}})}{\partial e_{\text{clean}}}
\end{equation}
where $L(\mathbf{x})$ is the logit loss, $e_{\text{ablated}}$ denotes predictions after ablation of the target edge, and the right side of Equation (1) represents the computed absolute attribution score \cite{syed_attribution_2023}. After ranking the scores, we keep the top-k edges for further processing. 

\subsection{Step 3: Subgraph Selection and Edge Pruning}
\label{ep}
From here, edges with low attribution scores are masked to produce a high-potential subgraph. The masking threshold balances sparsity against the retention of potentially cooperative but weakly attributed components (e.g. S-inhibition heads). We use the edges found by EAP to "jumpstart" the EP training process. EP proceeds by optimizing a binary mask $z \in [0, 1]^{N_{\text{edge}}}$ to minimize output divergence between the original and pruned graphs, under a targeted sparsity constraint: 
\begin{equation}
1 - \frac{|H|}{|G|} \ge c
\end{equation}
This step is performed via gradient-based optimization using clean and corrupted examples \cite{bhaskar_finding_2025}.

\section{Experiment}
\subsection{Task Description}
\label{sec:ioi}
The task being studied is defined by a set of prompts that elicit a clearly defined response from the model predictions. We study the Indirect Object Identification (IOI) task, which is in the general format of "\textit{When Dylan and Ryan went to the store, Dylan gave a popsicle to $\rightarrow$ Ryan}". We use \cite{wang_interpretability_2022}'s prompt templates to generate an IOI dataset of 200 randomly selected examples with lexical and syntactic diversity, each for training and validation. Our test split involved 36,084 examples as per \cite{bhaskar_finding_2025}. 
\subsection{Experimental Setup}
\label{exp}
We evaluate our Hybrid Attribution and Pruning (HAP) framework on the Indirect Object Identification (IOI) task using GPT-2 Small (117M). The attribution score threshold in EAP is set very low to preserve possibly cooperative edges that might score low individually. For EP, we use the hyperparameters as detailed in \cite{bhaskar_finding_2025}. All training runs were performed on one NVIDIA H100 GPU. We quantify circuit quality with faithfulness via KL divergence and logit difference between model predictions and circuit predictions, and report standard metrics such as accuracy and runtime.

\section{Results}
\label{results}
\subsection{HAP vs Existing Methods}
\label{hap_results}
To compare the performance between different models, we leverage manually discovered circuits in \cite{wang_interpretability_2022} as a reference to calculate the accuracy of circuits recovered by automatic methods. As shown in Table \ref{gpt2}, HAP outperforms EAP in accuracy while having only slightly lower accuracy compared to EP. Similarly, circuits recovered by HAP are much more faithful to the full model compared to EAP (when comparing logit difference), while also maintaining similar faithfulness to EP circuits. It is shown in both KL divergence and logit difference metrics that HAP circuits are only slightly less faithful than EP circuits. 

When GPU and target sparsity are controlled, HAP is at least 46\% faster than EP while maintaining high accuracy and faithfulness to the full model. This shows that HAP can be a valuable framework for reducing the computational cost of circuit discovery, possibly enabling scalability to larger models.  
\begin{table}
    \centering
    \begin{tabular}{cccccc}\toprule
  \textbf{Algorithm}&\textbf{Sparsity}& \multicolumn{4}{c}{GPT-2 Small}\\\midrule
 & & \textbf{Accuracy} $\uparrow$& \textbf{Logit Diff} $\uparrow$&\textbf{KL} $\downarrow$& \textbf{Runtime} (s) $\downarrow$\\\midrule
 EAP & 94$\pm$0.5\%& 0.698&  3.13&\textendash& 
4\\
         EP &94$\pm$0.5\%&  0.772&  3.48 &0.190&  
2921\\
         HAP &94$\pm$0.5\%&  0.759&  3.42 &0.188&  1579\\ \bottomrule
    \end{tabular}
    
    \caption{Efficiency of HAP compared to existing works.}
    \label{gpt2}
\end{table}
\subsection{Case Study: S-inhibition Heads in IOI}
\label{s-heads}
To present the qualitative advantages of our hybrid framework, we present a case study on the IOI task in GPT-2 Small (see Section ~\ref{sec:ioi}). In IOI, the role of S-inhibition heads (or Subject-Inhibition Heads) is cooperative: they suppress the Name Mover Heads from incorrectly flagging the subject of a sentence due to their proximity to the verb. Thus S-Inhibition Heads, although critical for accurate task performance, are difficult to detect due to the low individual importance assigned by methods like \cite{syed_attribution_2023} at high sparsity.

\begin{figure}
    \centering
    \includegraphics[width=1\linewidth]{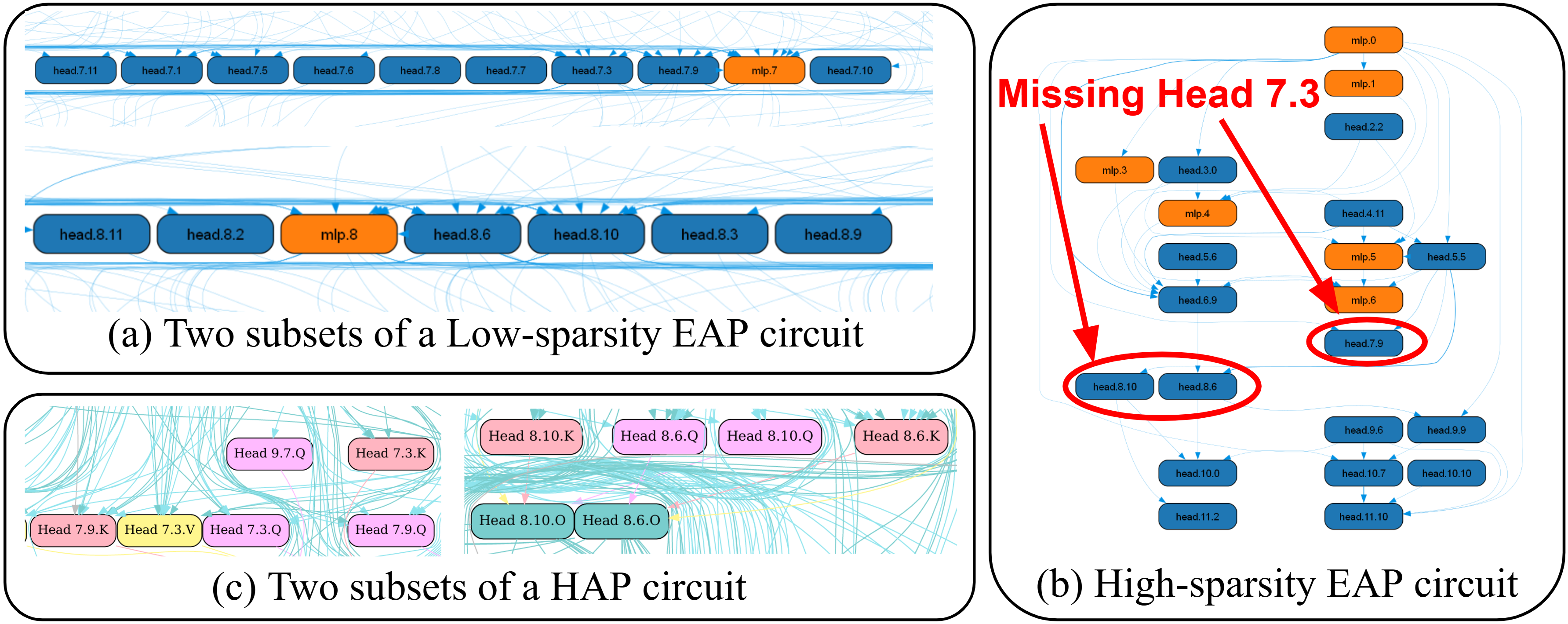}
    \caption{Recovered IOI circuits. While EAP on its own is unable to recover all S-Inhibition Heads at high sparsity, HAP preserves S-Inhibition Heads because it only uses EAP at low sparsity.}
    \label{fig:graphs}
\end{figure}

We found that existing methods do not recover the complete circuit. For example, EAP falls short since S-inhibition heads do not receive high attribution scores, causing them to be undervalued as shown in Figure \ref{fig:graphs}B. 
In contrast, HAP successfully captures the complete, functional circuit. By first using EAP to define a constrained search space with a generous threshold, we created a "safe zone" that retains these S-inhibition heads despite their low individual scores. The subsequent EP algorithm, operating on this focused and less noisy subgraph, correctly identifies their cooperative importance. As shown in Figure \ref{fig:graphs}C, the S-inhibition heads 7.3, 7.9, 8.6, and 8.10 are all preserved by HAP. This serves as qualitative evidence that our method is not only efficient but also preserves cooperative components that are overlooked by prior approaches. 
\section{Limitations}
\label{limit}
Our experiments are conducted exclusively on the IOI task with the GPT-2 Small model. Although this task is a well-established benchmark for mechanistic interpretability, further evaluation on a broader set of models and tasks is necessary to assess the generality, robustness, and scalability of HAP.
Furthermore, the current implementation has not optimized the threshold to select edges during the EAP stage, which will require future hyperparameter tuning. We also acknowledge that variations in the generated training dataset may result in minor performance differences across different runs.  

\section{Conclusion}
\label{concl}
We introduce HAP, a hybrid framework that resolves the longstanding speed-faithfulness tradeoff in circuit discovery by strategically sequencing EAP, a fast and approximate algorithm, with EP, a fine-grained and precise one. Our experiments show this approach is not only 46\% faster than EP while maintaining comparable faithfulness, but is also qualitatively superior. As demonstrated in our IOI case study, HAP successfully preserves the S-inhibition heads that attribution methods fail to recover in isolation. The results challenge the notion that the speed-faithfulness trade-off is fundamental and provide a simple framework to scale up future mechanistic interpretability research to interpret larger models.

\bibliographystyle{plainnat}
\bibliography{natbib}

\newpage
\appendix
\section{IOI Dataset Generation}
In Table \ref{tab:table2}, we provide the full set of IOI templates from \cite{wang_interpretability_2022} used to generate our dataset described in Section \ref{sec:ioi}. Names were sampled from a list of 100 common English first names, while places and objects were selected from a curated set of 20 frequent options. 
\begin{table}[h!]
\centering
\begin{tabular}{|p{14cm}|}
\hline
\textbf{IOI prompt templates}\\
\hline
Then, [B] and [A] went to the [PLACE]. [B] gave a [OBJECT] to [A] \\
\hline
Then, [B] and [A] had a lot of fun at the [PLACE]. [B] gave a [OBJECT] to [A] \\
\hline
Then, [B] and [A] were working at the [PLACE]. [B] decided to give a [OBJECT] to [A] \\
\hline
Then, [B] and [A] were thinking about going to the [PLACE]. [B] wanted to give a [OBJECT] to [A] \\
\hline
Then, [B] and [A] had a long argument, and afterwards [B] said to [A] \\
\hline
After [B] and [A] went to the [PLACE], [B] gave a [OBJECT] to [A] \\
\hline
When [B] and [A] got a [OBJECT] at the [PLACE], [B] decided to give it to [A] \\
\hline
When [B] and [A] got a [OBJECT] at the [PLACE], [B] decided to give the [OBJECT] to [A] \\
\hline
While [B] and [A] were working at the [PLACE], [B] gave a [OBJECT] to [A] \\
\hline
While [B] and [A] were commuting to the [PLACE], [B] gave a [OBJECT] to [A] \\
\hline
After the lunch, [B] and [A] went to the [PLACE]. [B] gave a [OBJECT] to [A] \\
\hline
Afterwards, [B] and [A] went to the [PLACE]. [B] gave a [OBJECT] to [A] \\
\hline
Then, [B] and [A] had a long argument. Afterwards [B] said to [A] \\
\hline
The [PLACE] [B] and [A] went to had a [OBJECT]. [B] gave it to [A] \\
\hline
Friends [B] and [A] found a [OBJECT] at the [PLACE]. [B] gave it to [A] \\
\hline
\end{tabular}
\caption{Templates used in the IOI dataset. The table displays templates following the “BABA” pattern; templates with the “ABBA” pattern were also employed but are omitted here for conciseness.}
\label{tab:table2}
\end{table}
\section{Connecting EAP to EP} 
\begin{figure}[H]
    \centering
    \includegraphics[width=1\linewidth]{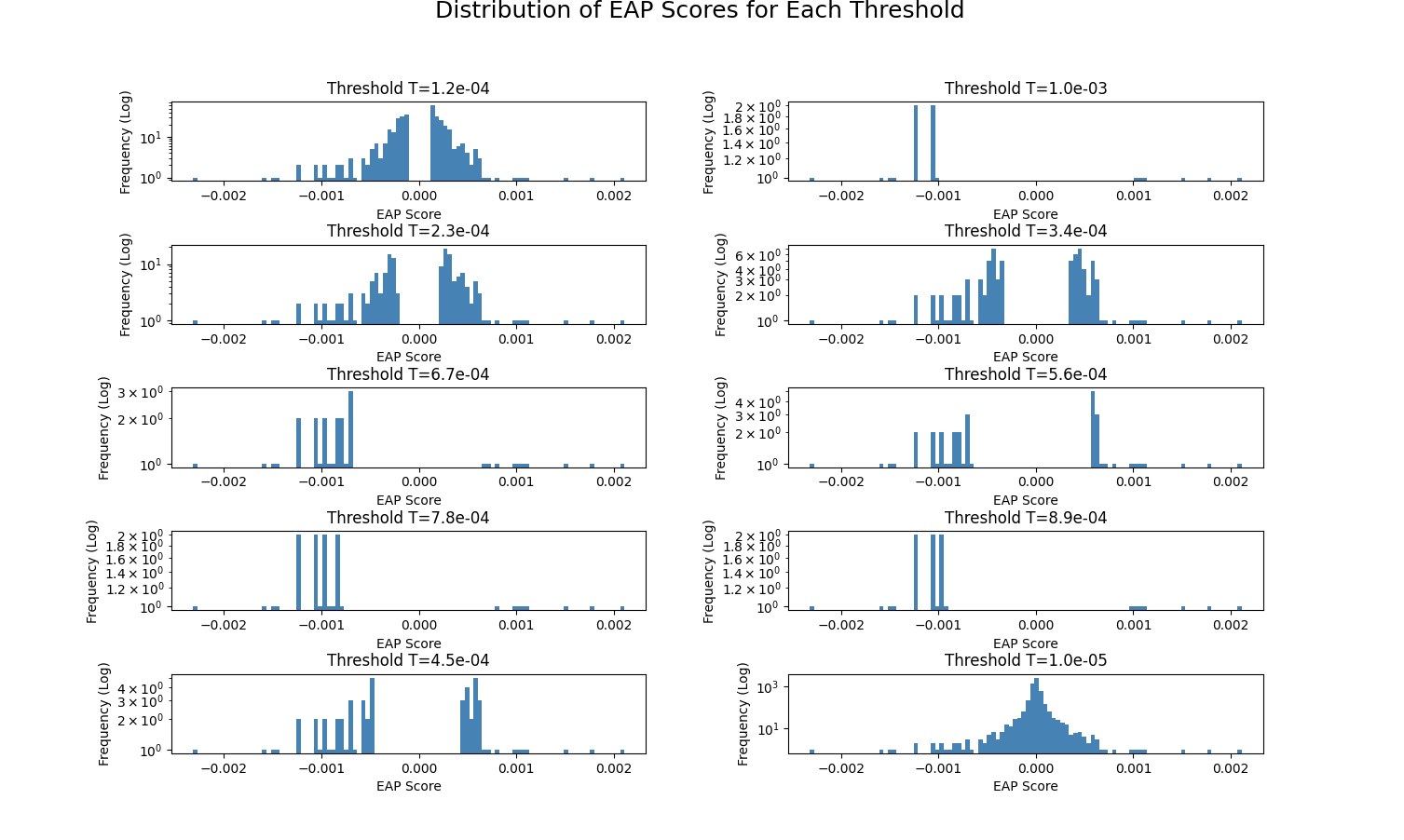}
    \caption{Attribution score distribution over different EAP thresholds.}
    \label{fig:supp}
\end{figure}
To map the high-potential edges identified by EAP to the binary masks $z$ of EP, we first show that EAP attribution scores are generally normally distributed (Figure \ref{fig:supp}). Then, we normalize the output attribution scores to a range $\in[-1,1]$. To integrate the normalized attribution scores into the binary masks $z$ of EP, we create the initial $log \space \alpha$ tensor. We then modify the EP initialization by changing the relevant mask parameters using the computed $log \space \alpha$ tensor. EP then undergoes training.

\end{document}